# Euclidean Auto Calibration of Camera Networks: Baseline Constraint Removes Scale Ambiguity

Kiran Kumar Vupparaboina, Kamala Raghavan and Soumya Jana

*Abstract*—Metric auto calibration of a camera network from multiple views has been reported by several authors. Resulting 3D reconstruction recovers shape faithfully, but not scale. However, preservation of scale becomes critical in applications, such as multi-party telepresence, where multiple 3D scenes need to be fused into a single coordinate system. In this context, we propose a camera network configuration that includes a stereo pair with known baseline separation, and analytically demonstrate Euclidean auto calibration of such network under mild conditions. Further, we experimentally validate our theory using a four-camera network. Importantly, our method not only recovers scale, but also compares favorably with the well known Zhang and Pollefeys methods in terms of shape recovery.

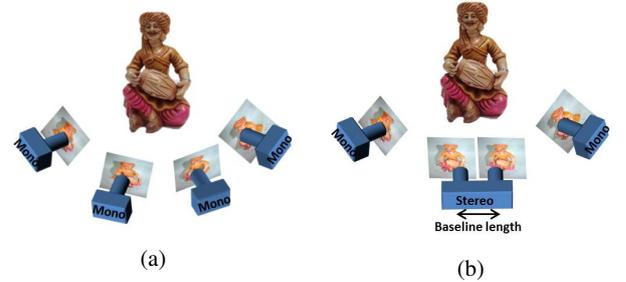

Fig. 1: Camera network: (a) classical; (b) proposed network including a stereo pair with known baseline length.

## I. INTRODUCTION

In a variety of applications, including telepresence [1], remote surgery [2], and virtual edutainment [3], [4], one aspires to reconstruct the three-dimensional (3D) likeness of objects at a remote location. This is often attempted based on views captured by a multitude of cameras [5], [6]. As a first step, the cameras are calibrated, i.e., the intrinsic parameters such as focal length, as well as the extrinsic parameters such as camera position and orientation are estimated. Traditionally, extraneous objects, such as chessboard patterns, checkered cubes and laser pointers, have been used for calibrating both single camera [7], [8], and network of cameras [1], [9], [10]. In contrast, scene-based auto calibration appears attractive in various scenarios, such as telepresence and remote classroom, where an adaptive camera network dynamically maintains the image quality, pre-calibrated cameras could be restrictive, and extraneous objects may not be introduced [11]. Further, applications such as multi-party telepresence, requiring fusion of multiple scenes, additionally demand preservation of scale. In this backdrop, we propose a Euclidean auto calibration method enabling faithful 3D reconstruction accurate to scale.

Traditionally, a multicamera network, depicted in Fig. 1a, consists of a number of monocular cameras. For such networks, various authors report metric auto calibration, which allows faithful recovery of shape, but not scale [5], [6], [11], [12]. Such a technique would typically obtain point correspondences from the multiple views at hand (practically, often using SIFT or SURF algorithms [13], [14], while removing false correspondences using RANSAC algorithm [5]), and then make use of either a factorization method [16], or Kruppa equations [17]. Adopting the former approach, Han and Kanade exploited inherent structural constraints, and faithfully recovered shape up to scale ambiguity [18], [19]. On the contrary, Kruppa equations, relating intrinsic parameters to the absolute conic, yield those parameters for a pair of cameras up to a scale, thereby introducing projective ambiguity in 3D reconstruction [20]. By fixing the plane at infinity, metric calibration was obtained for invariant and later for slowly varying intrinsic parameters [21], [22], [23]. Metric calibration has also been obtained by imposing other related constraints [24]–[31]. Further, attempts have been made at combining factorization and Kruppa equations towards distributed auto calibration [32]. Subsequently, making use of efficient sparse bundle adjustment (SBA) [33], Furukawa and Ponce proposed a method for auto calibration and dense 3D reconstruction [34]. At the same time, metric auto calibration has been obtained in the stochastic framework [35], [36], [37].

In comparison, significantly less effort has been directed towards Euclidean auto calibration, required for recovery of both shape and scale. Lerma *et al.* considered a three-camera configuration, imposed baseline constraint on each of the three camera pairs (resulting in an inflexible structure), and reported photogrammetrically improved auto calibration [38]. However, questions on uniqueness of the solution and scalability to larger networks were not addressed. In contrast, building on our earlier work [39], we propose a configurable camera network including a stereo pair with known baseline separation, as depicted in Fig. 1b, and analytically establish its Euclidean auto calibration accurate to scale. Clearly, our analysis applies to the three-camera network of Lerma *et al.*, and hence, just one of their three baseline constraints suffices

K. K. Vupparaboina and S. Jana are with the Department of Electrical Engineering, Indian Institute of Technology Hyderabad, Andhra Pradesh, 502205 India e-mail: {ee11p012, jana}@iith.ac.in

K. Raghavan is with the Department of Electrical Engineering, Indian Institute of Technology Madras, Tamil Nadu, 600036 India e-mail: ee10b126@smail.iitm.ac.in

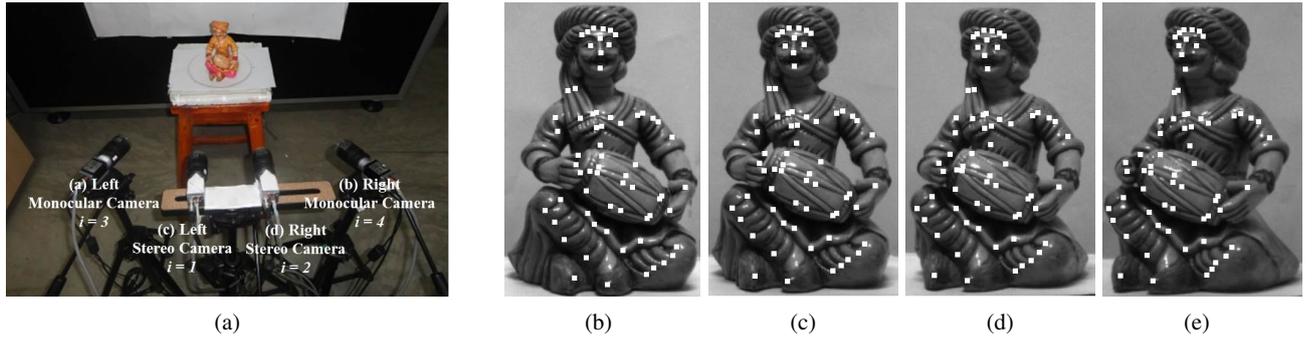

Fig. 2: (a) Proposed four-camera network including a stereo pair and two monocular cameras; (b), (c), (d), (e) respective images, with highlighted feature points, taken by left monocular, left stereo, right stereo, and right monocular cameras.

for Euclidean auto calibration. Subsequently, we validate our theory experimentally using an arbitrary four-camera network including a stereo pair. In particular, we demonstrate an accurate recovery of the scale, and the shape by directly comparing with the original object using a state-of-the-art lightfield display [40]. Significantly, the proposed method recovers scale from the mild baseline constraint, while still achieving reprojection errors comparable to that achieved by the well-known Zhang and Pollefeys methods [8],[23].

The rest of the paper is organized as follows. The proposed auto calibration problem is formulated as a multi-objective minimization in Sec. II, while the uniqueness condition is derived in Sec. III. Further, Sec. IV experimentally validates our theoretical results. Finally, Sec. V concludes the paper.

## II. Problem Formulation

Consider an $M$-camera network. Further, consider $N$ feature points, with homogeneous coordinates $\bar{X}_j = [X_j\ Y_j\ Z_j\ 1]^T$ ($j = 1, 2, ..., N$), each visible by all the $M$ cameras [18]. Denote by $\bar{x}_{ij} = [x_{1j}\ y_{1j}\ 1]^T$ the projection of $\bar{X}_j$ on the $i$-th image plane (in local coordinates), $i = 1, 2, ..., M$. Auto calibration consists in estimating the intrinsic and the extrinsic parameters of the $M$ cameras from $\bar{x}_{ij}$ ($i = 1, 2, ..., M$, $j = 1, 2, ..., N$). Assume that the first two cameras ($i = 1, 2$) form a stereo pair [41] with parallel optical axis and given baseline separation $l$. A typical deployment with $M = 4$ is shown in Fig. 2a. We shall show that unambiguous auto calibration is possible if $l$ is accurately known. This in turn enables true-to-scale recovery of 3D point cloud $\{\bar{X}_j\}$.

### A. Image formation

Ignoring lens distortion or assuming it is corrected already, image $\bar{x}_{ij}$ of $\bar{X}_j$ ($j = 1, 2, ..., N$) by the $i$-th ($i = 1, 2, ..., M$) camera is given by [5]

$$s_{ij}\begin{bmatrix} x_{ij} \\ y_{ij} \\ 1 \end{bmatrix} = \begin{bmatrix} f_i & 0 & 0 \\ 0 & f_i & 0 \\ 0 & 0 & 1 \end{bmatrix}\begin{bmatrix} r_1^i & r_2^i & r_3^i & | & t_X^i \\ r_4^i & r_5^i & r_6^i & | & t_Y^i \\ r_7^i & r_8^i & r_9^i & | & t_Z^i \end{bmatrix}\begin{bmatrix} X_j \\ Y_j \\ Z_j \\ 1 \end{bmatrix}, \quad (1)$$

where $s_{ij}$ denotes a scale factor. In the right hand side of (1), the first matrix of intrinsic parameters including focal length $f_i$ is diagonal under the assumption that the origin of the image plane lies at its intersection with the principal axis, and the skew factor is zero. The second matrix of extrinsic parameters consists of a $3 \times 3$ unitary rotation parameter matrix concatenated with a $3 \times 1$ translation parameter vector as shown. Our task is to determine all other quantities based on the $MN$ images $\{\bar{x}_{ij}\}$ of the $N$ feature points.

Recall that cameras '1' and '2' constitute the stereo pair, and the rest are indexed '3',...,'$M$'. Now fix the origin of the world coordinate system at the center of Camera 1, and take its principal axis as the $Z$-axis. Thus the center of Camera 2 has the location $[l\ 0\ 0]^T$. Further, for simplicity, assume that the sensor arrays in Cameras 1 and 2 are shifted versions of each other such that both rotation matrices are identity(see, e.g., [41]). Accordingly, for $i = 1, 2$, (1) yields

$$X_j = \frac{x_{1j}y_{2j}l}{x_{1j}y_{2j} - x_{2j}y_{1j}} \quad (2) \qquad f_2 = \frac{y_{2j}}{y_{1j}}f_1 \quad (4)$$

$$Y_j = \frac{y_{1j}y_{2j}l}{x_{1j}y_{2j} - x_{2j}y_{1j}} \quad (3) \qquad Z_j = \frac{f_1 X_j}{x_{1j}}. \quad (5)$$

In view of (2)–(5), $f_1$ remains the only unknown quantity relating to the stereo pair ($i = 1, 2$). Here (4) forces the ratio $\frac{y_{2j}}{y_{1j}}$ to be the same for every $j = 1, 2, ..., N$. In practice, we shall take average ratio in view of possible errors.

Further, using (5) in (1) for $i \geq 3$, we obtain

$$x_{ij} = X_j\frac{f_i r_1^i}{t_Z^i} + Y_j\frac{f_i r_2^i}{t_Z^i} + \frac{X_j}{x_{1j}}\frac{f_1 f_i r_3^i}{t_Z^i} + \frac{f_i t_X^i}{t_Z^i}$$
$$- x_{ij}X_j\frac{r_7^i}{t_Z^i} - x_{ij}Y_j\frac{r_8^i}{t_Z^i} - \frac{x_{ij}X_j}{x_{1j}}\frac{f_1 r_9^i}{t_Z^i} \quad (6)$$

$$y_{ij} = X_j\frac{f_i r_4^i}{t_Z^i} + Y_j\frac{f_i r_5^i}{t_Z^i} + \frac{X_j}{x_{1j}}\frac{f_1 f_i r_6^i}{t_Z^i} + \frac{f_i t_Y^i}{t_Z^i}$$
$$- y_{ij}X_j\frac{r_7^i}{t_Z^i} - y_{ij}Y_j\frac{r_8^i}{t_Z^i} - \frac{y_{ij}X_j}{x_{1j}}\frac{f_1 r_9^i}{t_Z^i}, \quad (7)$$

which gives rise to multi-objective constrained optimization.

### B. Multi-objective optimization

Stacking (6) and (7) ($j = 1, 2, ..., N$), we get respectively

$$A_x^i \bar{\alpha}^i = \bar{b}_x^i \quad (8) \qquad\qquad A_y^i \bar{\beta}^i = \bar{b}_y^i, \quad (9)$$

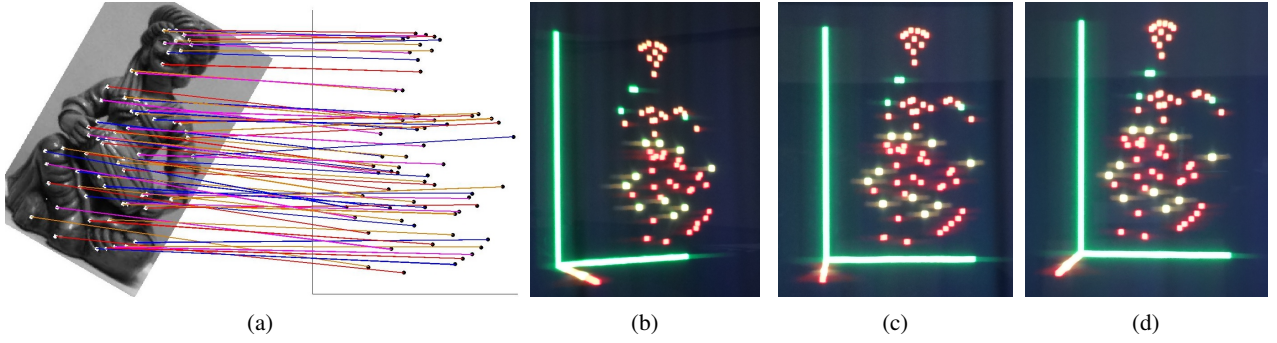

Fig. 3: 3D visualization of reconstructed point cloud: (a) orthographic projection with feature point correspondence with 2D perspective view; (b), (c), (d) rendering on 3D lightfield display [40]: images taken from approximate perspectives of the left monocular, left stereo, and right monocular cameras, respectively.

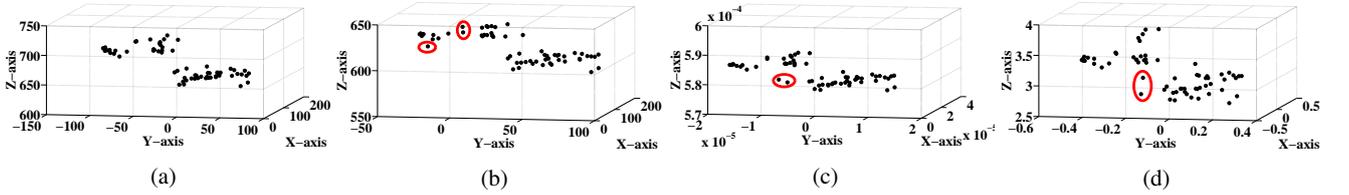

Fig. 4: 3D point cloud comparison: (a) proposed method, (b) Zhang method with BA, (c) Pollefeys method with BA, and (d) proposed method degraded by BA. Grossly mislocated feature points are circled in red.

where $\bar{b}_x^i = \begin{bmatrix} x_{i1} & \ldots & x_{iN} \end{bmatrix}^T$, $\bar{b}_y^i = \begin{bmatrix} y_{i1} & \ldots & y_{iN} \end{bmatrix}^T$,

$$\bar{\alpha}^i = \begin{bmatrix} \frac{f_i r_1^i}{t_Z^i} & \frac{f_i r_2^i}{t_Z^i} & \frac{f_i f_1 r_3^i}{t_Z^i} & \frac{f_i t_X^i}{t_Z^i} & \frac{r_7^i}{t_Z^i} & \frac{r_8^i}{t_Z^i} & \frac{f_1 r_9^i}{t_Z^i} \end{bmatrix}^T, \quad (10)$$

$$\bar{\beta}^i = \begin{bmatrix} \frac{f_i r_4^i}{t_Z^i} & \frac{f_i r_5^i}{t_Z^i} & \frac{f_i f_1 r_6^i}{t_Z^i} & \frac{f_i t_Y^i}{t_Z^i} & \frac{r_7^i}{t_Z^i} & \frac{r_8^i}{t_Z^i} & \frac{f_1 r_9^i}{t_Z^i} \end{bmatrix}^T, \quad (11)$$

$$A_x^i = \begin{bmatrix} X_1 & Y_1 & \frac{X_1}{x_{11}} & 1 & -x_{i1}X_1 & -x_{i1}Y_1 & -\frac{x_{i1}X_1}{x_{11}} \\ \vdots & & & \ddots & & & \vdots \\ X_N & Y_N & \frac{X_N}{x_{1N}} & 1 & -x_{iN}X_N & -x_{iN}Y_N & -\frac{x_{iN}X_N}{x_{1N}} \end{bmatrix} \quad (12)$$

and $A_y^i$ takes the same form as $A_x^i$ with each occurrence of $x_{ij}$ replaced by $y_{ij}$ ($j = 1, 2, ..., N$).

In practical error-prone scenarios, rather than (8) and (9), we propose to solve a multi-objective minimization problem:

$$\min \sum_{i=3}^{M} \lambda_i \|A_x^i \bar{\alpha}^i - \bar{b}_x^i\| + (1-\lambda_i)\|A_y^i \bar{\beta}^i - \bar{b}_y^i\|, \quad (13)$$

where $\lambda_i \in [0\ 1]$ determines the relative weights of reprojection error along $x$ and $y$ axes, $i = 1, 2, ..., M$.

## III. EUCLIDEAN AUTO CALIBRATION

Since components of $\bar{\alpha}^i$ and $\bar{\beta}^i$ are dependent (refer (10) and (11)), constituent least squares problems in (13) are nonlinear. Specifically, $f_1$ and $\{f_i, r_1^i, r_2^i, r_4^i, t_X^i, t_Y^i, t_Z^i\}_{i=3}^{M}$ constitute a set of $7(M-2)+1$ independent variables. To see this, it is enough to verify that $r_1^i$, $r_2^i$ and $r_4^i$ specify the $3 \times 3$ unitary rotation matrix in (1). First, write [7]

$$r_5^i = \frac{-r_1 r_2 r_4}{1 - r_1^2} \\ \pm \frac{\sqrt{1 - 2r_1^2 - r_2^2 - r_4^2 + r_1^2 r_2^2 + r_1^2 r_4^2 + r_2^2 r_4^2 + r_1^4}}{1 - r_1^2} \quad (14)$$

(dropping index '$i$' on the right hand side), and hence, using the unitarity property, obtain $r_3^i$, $r_6^i$, $r_7^i$, $r_8^i$ and $r_9^i$ up to respective signs. These sign ambiguities translate to eight options for the desired rotation matrix [39]. Thus, theoretically, one may need to solve up to a finite number $8^{M-2}$ of minimization problems. Practically, a divide-and-conquer method could reduce the number of problems to solve.

To ensure unique solution, we desire the system to be overdetermined. Since we have $2N$ equations for the $i$-th ($i = 3, ..., M$) camera in the form of (8) and (9), i.e., $2(M-2)N$ equations altogether, and $7(M-2)+1$ variables, we require

$$(M-2)(2N-7) \geq 1. \quad (15)$$

From (15), we find the following conditions to be both necessary and sufficient: (i) number of cameras $M \geq 3$, and (ii) number of feature points $N \geq 4$. Finally, upon finding the independent variables minimizing (13), the accurate-to-scale 3D point cloud $\{\bar{X}_j\}$ can be obtained from (2), (3) and (5).

## IV. EXPERIMENTAL CORROBORATION

At this point, we set up an experimental four-camera network ($M = 4$), as shown in Fig. 2a, using Basler Ace 1300-30gc GigE cameras [42], fitted with Goyo Optics GMDN24012C lenses [43]. The baseline length for the stereo camera pair ($i = 1, 2$) is set at $l = 125$mm. Next we perform autocalibration as proposed, based on images of $N = 57$ feature points, indicated by white dots, in each of the four views shown in Figs. 2b–2e. We now demonstarte accurate Euclidean autocalibration, by establishing accurate recovery of the shape (geometry) and the scale of the reconstructed 3D cloud of the aforementioned feature points.

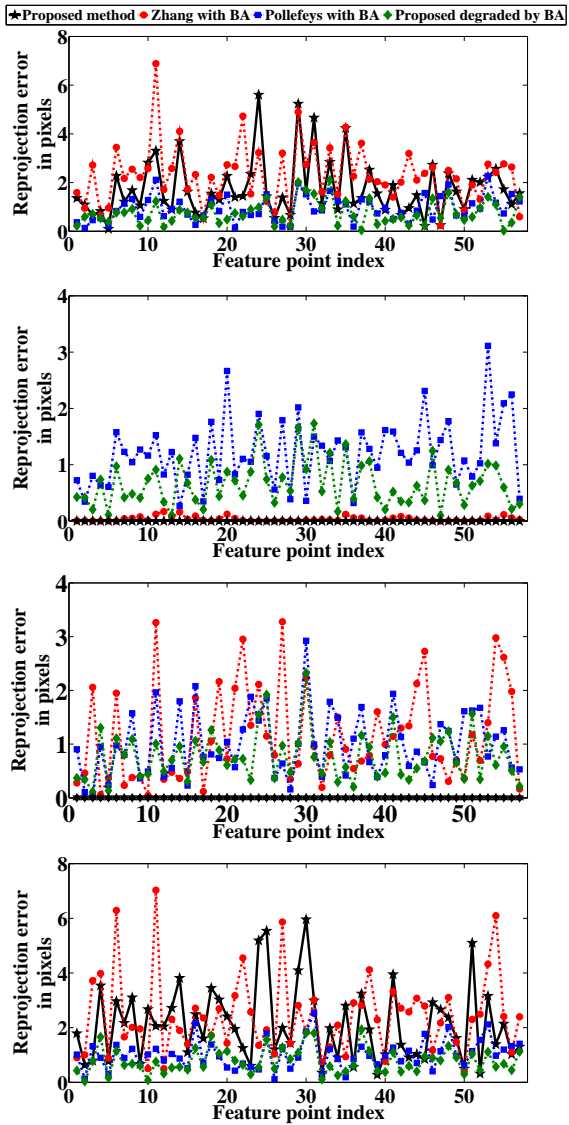

Fig. 5: Comparison of reprojection error: (a) left monocular (b) left stereo, (c) right stereo, and (d) right monocular cameras.

To this end, in Fig. 3a, we first compare a classical scatter plot (an orthographic view) of reconstructed point cloud, along with point correspondences in the left stereo view ($i = 1$, a perspective projection). Despite certain similarity, comparison between such orthographic and perspective views remains problematic. Fortunately, a state-of-the-art lightfield display (Holovizio 721RC [40]), capable of true 3D rendering, allows direct visual comparison among collections of 3D points. Consequently, 2D images (perspective views) of the true-3D-rendered point cloud can directly be compared with corresponding images of the original. In Figs. 3b, 3c and 3d, respectively, we present views of the reconstructed point cloud from the approximate perspectives of the left monocular, the left stereo and the right monocular cameras. Comparing with the corresponding original views shown in Figs. 2b, 2c and 2e, a high degree of likeness in geometry (shape) is observed.

Next we compare the scales and shapes recovered using the proposed and competing algorithms. Our method yields a cloud, shown in Fig. 4a, that fits inside a rectangular parallopiped of dimensions 98mm $\times$ 168mm $\times$ 75mm ($X \times Y \times Z$), which closely match physical measurements. The corresponding shape, authenticated visually in Figs. 3b, 3c and 3d, is now taken as a reference. In comparison, the Zhang method with bundle adjustment (BA) [8],[44], shown in Fig. 4b, finds the point cloud inside a volume 72mm $\times$ 124mm $\times$ 46mm, which considerably deviates from the original scale. This happens because the accurate scale information, obtained during calibration of individual cameras using a chessboard with known dimension, is lost during BA. Also, the shape, with clearly misplaced features, matches the original only roughly, showing performance limits of BA. Further, the Pollefeys method with BA [23], shown in Fig. 4c, clearly ignores the scale, as expected from the theory. However, this method reconstructs the shape reasonably well, which indicates its efficacy, because we have only four as opposed to the cumtomarily large number of views.

For the sake of completeness, we now turn to comparing reprojection errors. To this end, in Fig. 5, we plot such error for each feature point and each of the four cameras in three cases, the proposed method, the Zhang method with BA [8], and the Pollefeys method with BA [23]. Notice that the reprojection errors for these three methods are comparable for monocular cameras, while the proposed method, as expected, incurs negligible errors in the stereo cameras. At this point, an attentive reader may ask: Would additional BA improve the proposed method, as it generally does the Zhang and Pollefeys methods? Interestingly, the answer is no. In fact, if subjected to additional BA, our method would need to violate the fundamental baseline constraint, and as depicted in Fig. 4d, would accordingly cause the reconstructed point cloud to lose the scale and the shape with major mislocations. However, notice in Fig. 5 that the proposed method, uopn degradation by additional BA as above, manages to obtain less reprojection error in general, by violating the fundamental baseline constraint. This clearly demonstrates that low reprojection error, by itself, is a poor indicator of callibration accuracy.

## V. Discussion

In this paper, we proposed a multicamera network that includes a stereo pair with known baseline separation, and demonstrated, both analytically and experimentally, Euclidean auto calibration of such networks. Our result provides a framework for true-to-scale 3D recovery of objects without interfering with those, thus opening up the possibility of adaptive and unobtrusive reconstruction of dynamic scenes. Of course, to realize a practical system, one would further require faithful surface reconstruction [34], and realistic rendering [45]. Here we hasten to add that ours remains one approach among several towards 3D reconstruction [46]–[49].

## REFERENCES

[1] H. Baker, D. Tanguay, I. Sobel, D. Gelb, M. Gross, W. Culbertson, T.Malzenbender, " The coliseum immersive teleconferencing system," *International Workshop on Immersive Telepresence*, France, 2002.

[2] R. M. Satava, "3-D vision technology applied to advanced minimally invasive surgery systems," *Surgical endoscopy*, vol. 7, no. 5, pp. 429–431, 1993.

[3] G. Jones, and M. Christal, "The future of virtual museums: On-line, immersive, 3d environments," *Created realities group*, vol. 4, 2002.

[4] Y. Shi, W. Xie, and G. Xu, "Smart remote classroom: Creating a revolutionary real-time interactive distance learning system," *Advances in Web-Based Learning*, pp. 130–141, Springer Berlin Heidelberg, 2002.

[5] R. Hartley, and A. Zisserman, *Multiple View Geometry in Computer Vision,* Cambridge University Press, Second Ed., 2004.

[6] H. Aghajan, and A. Cavallaro, (Eds.) *Multi-Camera Networks: Principles and Applications*, Academy Press, May 2009.

[7] R.Y. Tsai, "A Versatile Camera Calibration Technique for High-Accuracy 3D Machine Vision Metrology Using Off-the-Shelf TV Cameras and Lenses,"*IEEE Journal of Robotics and Automation*, vol. 3, no. 4, pp. 323–344, August 1987.

[8] Z. Zhang, "A Flexible New Technique for Camera Calibration," *IEEE Transactions on Pattern Analysis and Machine Intelligence (TPAMI)*, vol. 22, no. 11, November 2000.

[9] G. Kurillo, L. Zeyu, and R. Bajcsy, "Wide-area external multi-camera calibration using vision graphs and virtual calibration object," *ACM/IEEE International Conference on Distributed Smart Cameras*, pp 1–9, California, USA, September 2008.

[10] T. Svoboda, D. Martinec, and T. Pajdla, "A convenient multi-camera self-calibration for virtual environments," *PRESENCE: Teleoperators and Virtual Environments*, vol. 14, no. 4, pp. 407–422, August 2005.

[11] E. E. Hemayed, "A survey of camera self-calibration," *IEEE Conference on Advanced Video and Signal Based Surveillance*, pp. 351–357, Miami, FL, USA, 2003.

[12] A. Gruen, and T. S. Huang, (Eds.) *Calibration and Orientation of Camera in Computer Vision*, Springer, Berlin, 2001.

[13] D. G. Lowe, "Object recognition from local scale-invariant features," *IEEE international conference on Computer vision (ICCV)*, vol. 2, pp. 1150–1157, Kerkyra, Greece, 1999.

[14] H. Bay, A. Ess, T. Tuytelaars, and L. Van Gool, "SURF: Speeded Up Robust Features," *Computer Vision and Image Understanding (CVIU)*, vol. 110, no. 3, pp. 346–359, 2008.

[15] B. Triggs, "Autocalibration and the absolute quadric," *IEEE Conference on Computer Vision and Pattern Recognition (CVPR)*, pp. 609–614, San Juan, Puerto Rico, 1997.

[16] P. Strum, and B. Triggs, "A factorization based algorithm for multi-image projective structure and motion," *European Conference on Computer Vision (ECCV)*, pp. 709–720, Cambridge, UK, 1996.

[17] S. Maybank, and O.D. Faugeras, "A theory of self-calibration of a moving camera," *International Journal on Computer Vision*, vol. 8, no. 2, 123–151, 1992.

[18] M. Han, and T. kanade,"Creating 3D Models with Uncalibrated Cameras,"*IEEE Workshop on Applications of Computer Vision (WACV)*, pp. 178–185, Palm Springs, California, USA, 2000.

[19] R. Szeliski, and S. B. Kang, "Shape ambiguities in structure from motion," *IEEE Transactions on Pattern Analysis and Machine Intelligence (TPAMI)*, vol. 19, no. 5, pp. 506–512, 1997.

[20] R.I. Hartley, "Kruppas equations derived from the fundamental matrix," *IEEE Transactions on Pattern Analysis and Machine Intelligence (TPAMI)*, vol. 19, no. 2, pp. 133–135, February 1997.

[21] M. Pollefeys, and L. Van Gool, "A stratified approach to metric self-calibration," *IEEE Conference on Computer Vision and Pattern Recognition (CVPR)*, pp. 407–412, San Juan, Puerto Rico, 1997.

[22] A. Heyden, and K. Astrom, "Euclidean reconstruction from image sequences with varying and unknown focal length and principal point," *In IEEE Conference on Computer Vision and Pattern Recognition (CVPR)*, pp. 438–443, San Juan, Puerto Rico, 1997.

[23] M. Pollefeys, R. Koch, and L. Van Gool, "Self-Calibration and Metric Reconstruction in spite of Varying and Unknown Internal Camera Parameters," *International Journal of Computer Vision*, vol. 32, no. 1, pp. 7–25, 1999.

[24] T. Moons, L. Van Gool, M. Proesmans, and E. Pauwels, "Affine reconstruction from perspective image pairs with a relative object-camera translation in between," *IEEE Transactions on Pattern Analysis and Machine Intelligence (TPAMI)*, vol. 18, no. 1, pp. 77–83, January 1996.

[25] O. Faugeras, L. Quan, and P. Sturm, "Self-Calibration of a 1D Projective Camera and Its Application to the Self-Calibration of a 2D Projective Camera," *IEEE Transactions on Pattern Analysis and Machine Intelligence (TPAMI)*, vol. 22, no. 10, pp. 1179–1185, October 2000.

[26] F. Lv, T. Zhao, and R. Nevatia, "Self-calibration of a camera from video of a walking human," *International Conference on Pattern Recognition (ICPR)*, vol. 1, pp. 562–567, Quebec City, Canada, 2002.

[27] P. Sturm, and S. Maybank, "On plane based camera calibration: a general algorithm, singularities, applications," *IEEE Conference on Computer Vision and Pattern Recognition, (CVPR)*, pp. 432-437, Fort Collins, CO, USA, 1999.

[28] G. Xu, J. Terai, and H. Shum, "A linear algorithm for Camera Self-Calibration, Motion and Structure Recovery for Multi-Planar Scenes from Two Perspective Images," *IEEE Conference on Computer Vision and Pattern Recognition (CVPR)*, pp. 474–479, Hilton Head Island, SC, June 2000.

[29] D. Liebowitz, and A. Zisserman, "Combining Scene and Auto-calibration Constraints", *International Conference on Computer Vision (ICCV)*, pp. 293–300, Corfu, Greece, 1999.

[30] J. Mendelsohn, and K. Daniilidis, "Constrained selfcalibration," *IEEE Conference on Computer Vision and Pattern Recognition (CVPR)*, pp. 581–587, Fort Collins, CO, USA, 1999.

[31] G. Wang, Q. M. Jonathan Wu, and W. Zhang, "Kruppa equation based camera calibration from homography induced by remote plane," *Pattern Recognition Letters*, vol. 29, no. 16, pp. 2137–2144, 2008.

[32] D. Devarajan, R. J. Radke, and H. Chung, "Distributed metric calibration of ad hoc camera networks," *ACM Transactions on Sensor Networks (TOSN)*, vol. 2, no. 3, pp. 380-403, 2006.

[33] M. I. Lourakis, and A. A. Antonis, "SBA: A software package for generic sparse bundle adjustment," *ACM Transactions on Mathematical Software (TOMS)*, vol. 36, no. 1, 2009.

[34] Y. Furukawa, and J. Ponce, "Accurate camera calibration from multi-view stereo and bundle adjustment," *International Journal of Computer Vision*, vol. 84, no. 3, pp. 257–268, 2009.

[35] G. Qian, and R. Chellappa, "Bayesian self-calibration of a moving camera," *Computer Vision and Image Understanding*, vol. 95, no. 3, pp. 287–316, 2004.

[36] J. Civera, D. R. Bueno, A. J. Davison, and J M. M. Montiel, "Camera self-calibration for sequential bayesian structure from motion," *IEEE International Conference on Robotics and Automation (ICRA)*, pp. 403–408, Kobe, Japan, 2009.

[37] A. Choudhary, G. Sharma, S. Chaudhury, and S. Banerjee, "Distributed calibration of pan-tilt camera network using multi-layered belief propagation," *In IEEE Computer Society Conference on Computer Vision and Pattern Recognition Workshops (CVPRW)*, pp. 33–40, June 2010.

[38] J. L. Lerma, S. Navarro, M. Cabrelles, and A.E. Seguí, "Camera calibration with baseline distance constraints," *Photogrammetric Record*, vol. 25, no. 130, pp. 140–158, 2010.

[39] K. K. Vupparaboina, R. Kamala, and S. Jana, "Smart Camera Networks: An Analytical Framework for Auto Calibration without Ambiguity," *IEEE conference on Recent advancements in Intelligent Computational Systems (RAICS)*, pp. 310–315, Kerala, India, December 2013.

[40] http://www.holografika.com/Products/HoloVizio-721RC.html

[41] J. Chen, W. Mustafa, A. Siddig, and W. Kulesza, "Applying dithering to improve depth measurement using a sensor-shifted stereo camera," *Metrology and Measurement Systems*, vol. 7, no. 3, pp. 335–347, 2010.

[42] http://www.baslerweb.com/products/ace.html?model=167

[43] http://www.goyooptical.com/products/cctv/manual/GMDN24012C.pdf

[44] B. Triggs, P. F. McLauchlan, R. I. Hartley, and A. W. Fitzgibbon, "Bundle adjustment—a modern synthesis," *Vision algorithms: theory and practice*, pp. 298–372, Springer Berlin Heidelberg, 2000.

[45] A. Reche-Martinez, I. Martin, and G. Drettakis, "Volumetric reconstruction and interactive rendering of trees from photographs," *ACM Transactions on Graphics (TOG)*, vol. 23, no. 3, pp. 720–727, 2004.

[46] S. Banerjee, S. Dutta, P. K. Biswas, and P. Bhowmick, "A low-cost portable 3D laser scanning system with aptness from acquisition to visualization." *Digital Heritage International Congress (DigitalHeritage)*, vol. 1, pp. 185–188, Marseille, France, 2013.

[47] S. Izadi, D. Kim, O. Hilliges, D. Molyneaux, R. Newcombe, P. Kohli, J. Shotton, S. Hodges, and A. Fitzgibbon, "KinectFusion: real-time 3D reconstruction and interaction using a moving depth camera," *ACM Symp. on User Interface Software and Technology*, pp. 559–568, 2011.

[48] J. Geng, "Structured-light 3D surface imaging: a tutorial," *Advances in Optics and Photonics*, vol. 3, no. 2, pp. 128–160, 2011.

[49] K. N. Kutulakos, and S. M. Seitz, "A theory of shape by space carving," *International Journal of Computer Vision*, vol. 38, no. 3, pp. 199–218, 2000.